\documentclass[10pt,twocolumn,letterpaper]{article}

\usepackage{iccv}
\usepackage{times}
\usepackage{epsfig}
\usepackage{graphicx}
\usepackage{amsmath}
\usepackage{amssymb}

\usepackage{enumitem}

\def\etal{\emph{et al}.}

\def\eg{\emph{e.g.}}
\def\ie{\emph{i.e.}}

\usepackage{bbold}
\usepackage{bbm}
\usepackage{booktabs} % toprule
\usepackage{array}  % 
\usepackage[tight,normalsize,sf,SF]{subfigure}
\usepackage{multirow}
\usepackage{url}
\usepackage{color}
\usepackage{float}
\usepackage{gensymb}

\usepackage[pagebackref=true,breaklinks=true,letterpaper=true,colorlinks,bookmarks=false]{hyperref}
\usepackage{colortbl}
\def\mycolor{\cellcolor[rgb]{0.8275,0.8275,0.8275}}

\iccvfinalcopy % *** Uncomment this line for the final submission

\def\iccvPaperID{1926} % *** Enter the ICCV Paper ID here

\ificcvfinal\fi
\begin{document}

%%%%%%%%% TITLE
\title{View N-gram Network for 3D Object Retrieval}

\author{Xinwei He$^1$$^*$ \qquad Tengteng Huang$^1$\thanks{indicates equal contributions.} \qquad Song Bai$^2$  \qquad Xiang Bai$^1$\thanks{corresponding author.}\\
 $^1$Huazhong University of Science and Technology\\
 $^2$University of Oxford\\
\tt\small \{eriche.hust, songbai.site\}@gmail.com \qquad \tt\small\{huangtengtng, xbai\}@hust.edu.cn 
 }

\maketitle
\thispagestyle{empty}

%%%%%%%%% ABSTRACT
\begin{abstract}

How to aggregate multi-view representations of a 3D object into an informative and discriminative one remains a key challenge for multi-view 3D object retrieval. Existing methods either use view-wise pooling strategies which neglect the spatial information across different views or employ recurrent neural networks which may face the efficiency problem. To address these issues, we propose an effective and efficient framework called View N-gram Network (VNN). Inspired by n-gram models in natural language processing, VNN divides the view sequence into a set of visual n-grams, which involve overlapping consecutive view sub-sequences. By doing so, spatial information across multiple views is captured, which helps to learn a discriminative global embedding for each 3D object. Experiments on 3D shape retrieval benchmarks, including ModelNet10, ModelNet40 and ShapeNetCore55 datasets, demonstrate the superiority of our proposed method. 

\end{abstract}

%%%%%%%%% BODY TEXT
\section{Introduction}

3D object retrieval is an important topic in computer vision and has received a surge of research attention owing to its close relationship with various geometry related applications,~\eg,~VR/AR~\cite{1310084,kim2017anatomical,takacs2008outdoors}, medical imaging~\cite{keysers2003statistical, bergamasco20183d, zhou2019semi} and 3D printing~\cite{uccheddu20183d}. 
%A large body of works have been proposed to address the problem of 3D retrieval.
With the development of 3D model acquisition technology, large amounts of 3D models are available for free, \eg,~the large-scale repository ShapeNet~\cite{chang2015shapenet}. Thanks to the advances of data-driven based deep learning techniques, dramatic progresses have been achieved in this field. Nowadays, research trend has shifted from designing hand-crafted features~\cite{belongie2001shape,chen2003visual,gao20103d,kazhdan2003rotation} to learn 3D shape representations directly via deep architectures~\cite{su2015multi,qi2017pointnet,maturana2015voxnet,kanezaki2018rotationnet, bai2019re}. 

In general, learning deep 3D shape representation can be coarsely divided into two mainstreams, \ie, model-based and view-based methods. Model-based methods~\cite{qi2017pointnet, qi2017pointnet++,wu20153d, maturana2015voxnet, xie2017deepshape} learn 3D shape representations directly from the raw representation (\eg,~point cloud, voxel) of 3D shapes. View-based methods~\cite{su2015multi,bai2016gift,wang2017dominant,shi2015deeppano, yu2018multi} usually first represent a 3D object with a set of 2D view images, then extract features of each view image, and finally aggregate them into a compact 3D shape descriptor. Comparing with model-based methods, view-based methods are more flexible and can benefit from recent developments in 2D image analysis, for instance, well-established architectures. Besides, in the real-world scenario, view images for 3D objects are easier to obtain, hence more efficient. % In addition, view-based methods mostly perform better than model-based method~\cite{su2015multi,savva2016shrec16,savva2017shrec}. 

\begin{figure}[tb]
\includegraphics[width=0.45\textwidth]{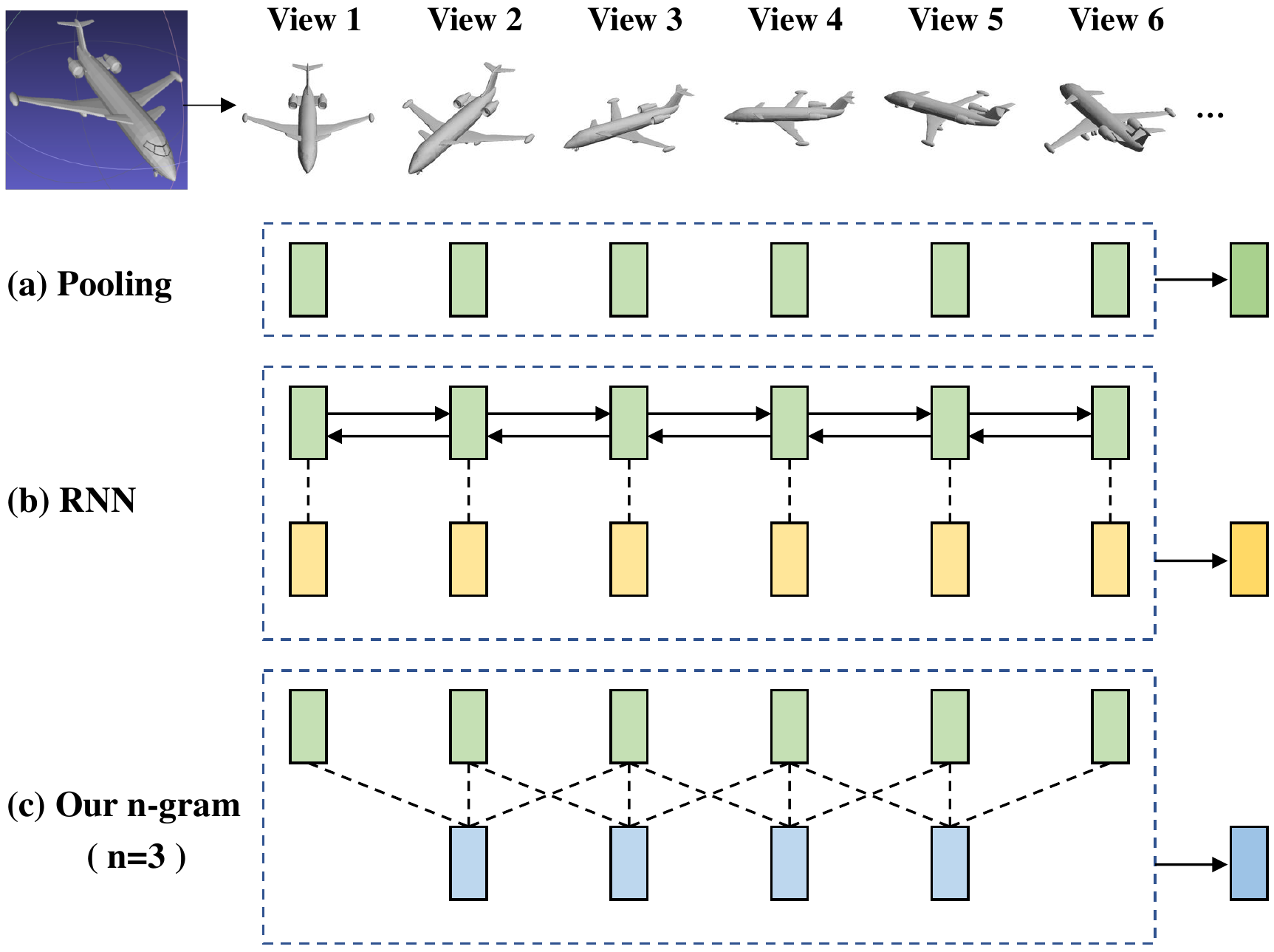}
\caption{Illustration of different aggregation strategies for multiple view images, including (a) view-wise pooling, (b) recurrent neural networks, and (c) the proposed visual n-grams.}
\label{fig:intro_model}
\vspace{-2ex}
\end{figure}

However, for multi-view based methods, one challenge is how to effectively aggregate the multiple view cues. To this end, existing works can be mainly categorized into two representative branches,~\ie,~view-wise pooling strategies~\cite{su2015multi,wang2017dominant} (see Fig.~\ref{fig:intro_model}(a)) and recurrent neural networks (RNN) based strategies~\cite{Dai2018Siamese, han2019SeqViews2SeqLabels} (see Fig.~\ref{fig:intro_model}(b)). Although great progresses have been achieved, these methods have certain limitations and cannot fully leverage the latent view embeddings. For instance, view-wise pooling strategies lose the spatial information across different views, while recurrent neural networks suffer from inefficiency due to the sequential working mechanism which requires much more computational resources~\cite{vaswani2017attention, li2016weighted}.

Considering these limitations, we propose an effective and efficient framework called View N-gram Network (VNN) based on the concept of n-gram (see Fig.~\ref{fig:intro_model}(c)) in natural language processing. In a typical n-gram model, a sentence is first decomposed into a sequence of n-grams, each of which has n consecutive words. Due to its capability of capturing word order in short context and its simplicity, n-gram has been widely used for various NLP tasks like language modeling~\cite{roark2007discriminative}, text classification~\cite{kim2014convolutional}, and machine translation~\cite{marino2006n}. Though its success has been demonstrated on learning text features, its effectiveness on learning 3D shape representations is unexplored.

However, n-gram also has intrinsic advantages for learning multi-view 3D shape representation. 
Specifically, for a 3D shape, we regard each view image as a ``word'' and divide the sequence of all the views into a set of overlapping consecutive sub-sequences which we call visual n-grams.
To this end, VNN utilizes an n-Gram Learning Unit (n-GLU) for efficient n-gram partitioning and intra-gram feature learning, thus the local spatial information among multiple views is well exploited. Compared with the RNN-based aggregation which works in a sequential manner, n-GLU is computationally more efficient as each visual n-gram only involves several consecutive views and can be fed in parallel to the network. Another merit of VNN is that the learned shape representation is rotation-invariant to a certain degree since the local adjacent relationship captured by visual n-grams is robust to rotation. Considering that visual n-grams of different sizes capture different scales of spatial information in the view sequence, we propose to combine the multi-scale representations learned with different n-gram sizes. Moreover, we further propose a parameterless attention model to selectively pack the partitioned n-gram features into a compact and global shape representation. 

To summarize, our main contributions are as follows: \begin{enumerate}[leftmargin=*, itemsep=0ex, partopsep=0pt,parsep=0ex]
 \item We present a novel framework named VNN to effectively model spatial information across the local context of the view sequence of each 3D shape. The proposed VNN firstly treats the rendered view sequence as a set of visual n-grams, and then computes rich n-gram features based on them, which produces more discriminative representations that are robust to rotation for 3D shape retrieval.
 \item To capture different scales of the local spatial information across the view sequence, we propose to combine the learning of distinct visual n-gram sizes, which can lead to further improvements.
 \item We design a parameterless attention model that efficiently and effectively aggregates learned visual n-gram features, which proves to be a better aggregation method than max-pooling for view-based 3D object retrieval task.
 \item Extensive experiments are conducted on both aligned and unaligned 3D shape benchmarks, and significant improvements are achieved over state-of-the-art methods.
\end{enumerate}

The rest of the paper is organized as follows. We briefly review related works in Section~\ref{sec:related_work}. Then, we elaborate the proposed VNN in Section~\ref{sec:proposed} and present the experimental evaluation in Section~\ref{sec:exp}. Conclusions are given in Section~\ref{sec:conclusion}. 

%-------------------------------------------------------------------------
\section{Related work}
\label{sec:related_work}
3D object retrieval has received increasing attention and great efforts have been devoted to constructing discriminative 3D shape descriptors. Early works mainly focus on designing handcrafted features to represent 3D shapes. Various types of 3D shape descriptors have been proposed, \eg, Light Field Descriptor (LFD)~\cite{chen2003visual}, Spherical Harmonic descriptor~\cite{kazhdan2003rotation}, and Heat Kernel Signatures~\cite{Bronstein2011Shape}.
% 3D shape retrieval has been extensively studied by many researchers. The key to 3D shape retrieval problem is the 3D shape descriptors. 
With the recent development of deep learning techniques, learning 3D shape representations by deep neural networks has become a hot topic in the 3D object retrieval field. Generally speaking, existing methods can be coarsely divided into two categories,~\ie~model-based methods and view-based methods. 

Model-based methods~\cite{xie2017deepshape,fang20153d} deal with the raw representations of 3D shapes directly,~\ie,~voxel, point cloud, and polygon mesh. Wu~\etal~\cite{wu20153d} propose 3D ShapeNets which  uses a Convolutional Deep Belief Network~(CDBN) to learn representations directly on the voxelized 3D objects. Similarly, Maturana and Scherer propose VoxNet~\cite{maturana2015voxnet}, which employs a 3D convolutional neural network and deals with 3D volumetric representations directly. Meanwhile, Qi~\etal~\cite{qi2016volumetric} propose Multi-Orientation Volumetric CNN (MO-VCNN) which aims at fusing the learned representations of 3D voxels from various orientations. However, methods based on the sparse volumetric representations are limited to the resolution of only $32^3$ due to the cubically increasing computational complexity and memory overhead. To address this problem, Wang~\etal~\cite{wang2017cnn} propose O-CNN, which is built upon a memory-efficient data structure named octree for the 3D object. Their method can process 3D shapes in the resolutions up to $256^3$. As for the representation of point clouds, PointNet~\cite{qi2017pointnet} is a pioneering work, which uses fully connected layers to embed the 3D coordinates into higher dimensional space and fuses them using max-pooling operation. Qi~\etal~\cite{qi2017pointnet++} further propose PointNet++ to extract local features with increasing contextual scales followed by a hierarchical aggregation mechanism. Klokov~\etal~\cite{klokov2017escape} propose Kd-Networks, which carry out computation based on the subdivision of point clouds using kd-trees. Overall, model-based methods are capable of exploiting the geometric information of 3D objects. 

View-based methods usually first project the raw 3D shapes to a panoramic view~\cite{shi2015deeppano,sfikas2018ensemble} or a set of 2D view images~\cite{su2015multi,wang2017dominant,kazhdan2003rotation,chen2018group, yu2018multi}. We mainly review the works which attempt to make rational use of multiple view images since they have a closer relation to our method. Bai~\etal~\cite{bai2016gift} propose a real-time 3D shape search engine based on the projective images. Su~\etal~~\cite{su2015multi} propose multi-view CNN, which uses a max-pooling operation to aggregate the multi-view representations outputted by a shared CNN. Dai~\etal~\cite{Dai2018Siamese} propose a siamese CNN-BiLSTM for 3D shape representation learning, where they use BiLSTM to capture features across different views of a 3D shape. In addition, Han~\etal~\cite{han2019SeqViews2SeqLabels} propose to use an RNN with attention to aggregate sequential views of each 3D object and promising results on several 3D shape retrieval benchmarks are obtained. Leng~\etal~\cite{leng2018learning} propose a score generation unit to evaluate the quality of the projected image and weight the view image features. 

Our proposed View N-gram Network (VNN) borrows the idea of n-gram to aggregate multi-view representations. The idea of n-gram has been widely applied to language models~\cite{kim2016character,jozefowicz2016exploring} and witnesses its success in text recognition~\cite{jaderberg2014deep, poznanski2016cnn}. For a given sequence, it can be sliced into a set of overlapping sub-sequences consisting of n consecutive words or characters, which are of great importance for exploring the pattern of sequences. However, there is no attempt to adapt the spirit of n-gram for 3D shape related tasks in the literature. In this paper, we divide the multi-view images into a set of small overlapping subgroups (hence we name it view N-gram), and perform view feature enhancement based on each subgroup. Finally, an attention mechanism is adopted to aggregate the enhanced features. We will detail our method in the following section.  

\section{View n-gram network}
\label{sec:proposed}
Given a 3D object $H$, we first render it into a set of 2D greyscale images $V_{H} = \{v_1, v_2, ..., v_{|V|}\}$, where $v_j$ denotes the $j$-th view image and $|V|$ represents the number of view images. Our goal is to learn a robust and discriminative representation for $H$ under the multi-view setting.

\begin{figure*}[tb]
\centering
\includegraphics[width=0.9\textwidth]{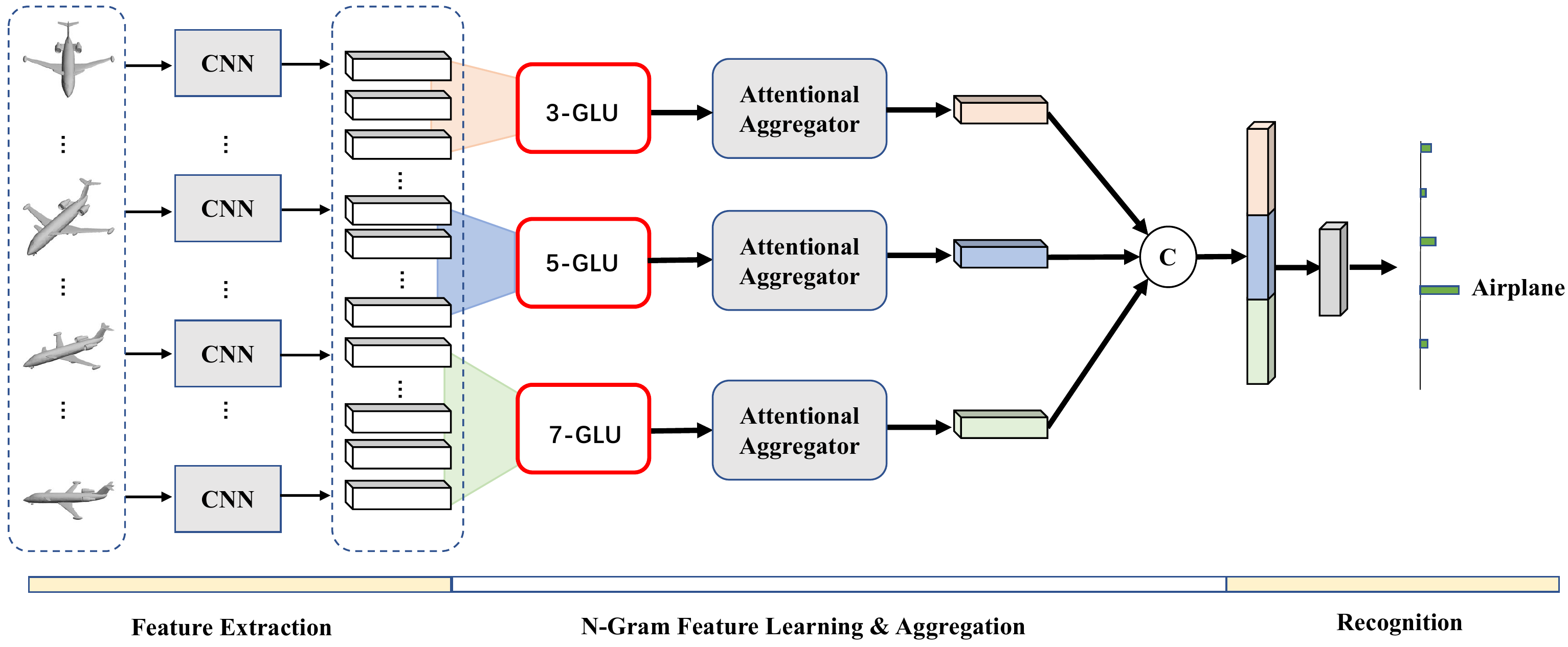}
\caption{The pipeline of View N-gram Network. A shared CNN is used to extract features for each view images of the 3D object. The raw feature sequence is enhanced by modeling the dependency of consecutive view features in a convolutional manner by our GLU module and then aggregated into a global descriptor. Three parallel branches are employed to exploit spatial information in different scales, resulting in a more discriminative representation for the 3D shape.}
\label{fig:fig_arch}
\vspace{-2ex}
\end{figure*}

As Fig.~\ref{fig:fig_arch} shows, the pipeline of the proposed View N-gram Network can be divided into 3 stages,~\ie,~feature extraction stage, n-gram feature learning and aggregation stage, and recognition stage. The first stage uses a shared convolutional neural network (CNN) that extracts features for each view, which is detailed in subsection~\ref{sec:view feature extraction}. The second stage is the core part of our framework, which is a multi-branch network with each branch consisting of an n-Gram Learning Unit~(n-GLU) (see subsection~\ref{sec:n-gram feature learning}) and a parameterless attentional feature aggregator (see subsection~\ref{sec:attentional feature aggregation}) to learn and aggregate n-gram features for particular visual n-gram size. The recognition stage and other supplementary details are given in subsection~\ref{sec:multi scale n-gram fusion}.

%-------------------------------------------------------------------------

\subsection{View feature extraction}
\label{sec:view feature extraction}
To extract the view features, a shared CNN is used. For each view image $v_i\in V_{H}$, the output of CNN is a $D$-dimensional feature $f_i\in\mathbb{R}^{D}$. Then, each 3D object can be represented as the multi-view embedding matrix $\mathbf{F} = [\mathbf{f}_1,\mathbf{f}_2, \mathbf{f}_3, ..., \mathbf{f}_{|V|}]^T\in \mathbb{R}^{|V| \times D}$ by concatenating $\mathbf{f}_i$ ($1\leq i\leq |V|$) according to the rendered order.

Note that any off-the-shelf convolutional neural network (\eg,~AlexNet~\cite{krizhevsky2012imagenet}, GoogLeNet~\cite{szegedy2015going}, ResNet~\cite{he2016deep}) can be used as the view feature extractor. In our work, we employ VGG-11 (also named VGG-A) with batch normalization~\cite{simonyan2014very} pre-trained on ImageNet~\cite{russakovsky2015imagenet} as our backbone. The original VGG-A has 11 layers, which consists of 8 convolutional layers (\textit{conv 1-8}) and 3 fully connected layers (\textit{fc 9-11}). In our experiments, we pre-train it and construct the feature extractor by removing the last two fully connected layers of the VGG-A. In this case, $D$ is 4096.

%-------------------------------------------------------------------------
\subsection{N-gram learning unit}
\label{sec:n-gram feature learning}

N-gram is a basic concept in natural language processing, and it has been widely used for language modeling. Let $S=(w_1, w_2, ..., w_m)$ be a sentence composed of $m$ words. An n-gram is defined as a substring consisting of n consecutive words $(w_i,w_{i+1}, ..., w_{i+n-1})$ from $S$. N-gram can adaptively model the temporal dependency of the $n$ consecutive words~\cite{li2016weighted}. 

As suggested in~\cite{han2019SeqViews2SeqLabels}, the spatial relationships among the view images play an important role in multi-view 3D shape analysis. Therefore, we propose to model the spatial dependency of view images in the form of n-gram inspired by its success in modeling temporal dependency. We regard each view image of a 3D object as one ``word'', then a sequence of $|V|$ rendered view images can be analogized to a sentence of $|V|$ words. Similarly, we can further decompose the view sequence into a set of $|V|+1-n$ n-grams, each of which is composed of $n$ consecutive view images (we call it visual n-gram). Each visual n-gram depicts a certain pattern of its corresponding 3D shape. Intuitively, 3D shapes from the same category should share similar n-gram patterns while those from different categories should differ in their n-gram patterns. Therefore, it would be beneficial for understanding a 3D shape to learn the n-gram patterns by capturing the local spatial dependency among consecutive view images. To this end, we propose a novel module named n-Gram Learning Unit~(n-GLU). And the mechanism works as follows. 

\begin{figure}[tb]
\includegraphics[width=0.45\textwidth]{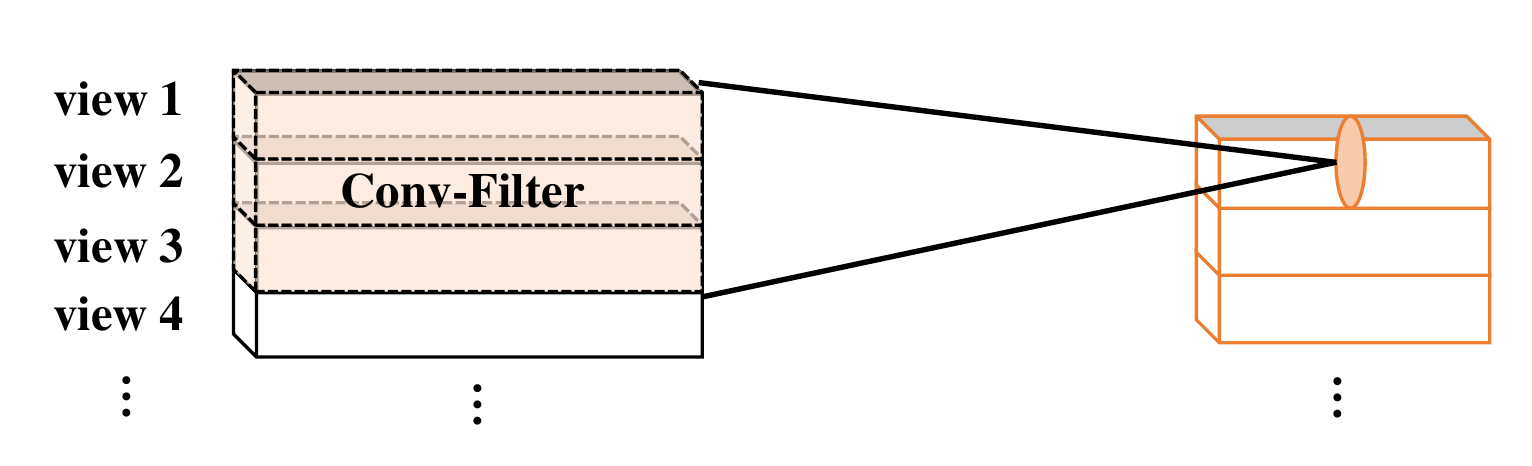}
\caption{A diagram of the GLU model used in our method. Here we choose the view n-gram size to be 3 for illustration purpose,~\ie,~3-GLU. In essence, GLU is a 2D convolutional filter, which convolves with the multi-view feature embedding matrix and produces visual n-gram features.}
\label{fig:GLU}
\vspace{-2ex}
\end{figure}

Recall that the multi-view representation for a 3D shape is denoted as an embedding matrix $\mathbf{F} \in \mathbb{R}^{|V| \times D}$ arranged in the rendering order. Similar to an n-gram based sentence classification network~\cite{kim2014convolutional}, we adopt a sliding window strategy of size $n \times D$ over $\mathbf{F}$ for the partition of n-grams as illustrated in Fig.~\ref{fig:GLU}. In particular, for each visual n-gram which corresponds to local consecutive n images of the sequence, we compute the enhanced visual n-gram features by using 2D convolution filter of size $ D' \times D \times n \times 1$, where $D'$ is the dimension of the output features enhanced by the visual n-gram. It is straightforward that the enhanced $D'$-dimensional representation has encoded the local spatial information of the corresponding visual n-grams. Since there are $|V|-n+1$ visual n-grams for each 3D object, therefore, the final enhanced compact representation $\mathbf{G}$ is of size $(|V|-n+1) \times D'$ for each 3D shape. In our experiments, $D'$ is set to 512.
% and the kernel size is $n\times1$. As a result, the raw multi-view representation $F$ is mapped to an enhanced representation $\mathbf{G} \in \mathbb{R}^{(|V|-n+1) \times D'}$ composed of $|V|-n+1$ n-grams. In our experiment, $D'$ is set to 512. 
% with local short range spatial information incroportated.  
%\textcolor{red}{For each n-gram, the view images involved in are mapped to a compact representation for corresponding n-gram view image (??????????????????????). For efficiency, we employ a 2D convolution to simulate this process. Concretely, the 2D convolution filter can be formulated as $ D' \times D \times n \times 1$, where $D'$ is the dimension of generated n-gram view image, and the kernel size is $n\times1$. As a result, the raw multi-view representation $F$ is mapped to an enhanced representation $\mathbf{G} \in \mathbb{R}^{(|V|-n+1) \times D'}$ composed of $|V|-n+1$ n-grams. In our experiment, $D'$ is set to 512}. \textcolor{blue}{The sliding window extracts features from all the view images inside it and output a compact representation for each visual n-gram.}
% We follow the work from \cite{kim2014convolutional} to design our \textbf{n-GLU} for multi-view feature enhancement. Concretely, 

N-gram learning unit possesses two desirable advantages for robust 3D representations. First, n-GLU can adaptively learn typical patterns for 3D shapes by exploring the spatial relation of several local consecutive views. Different from RNN which aims at modeling the long-range dependency among all the views, the proposed method can better capture local and fine-grained patterns. 
%Therefore, more discriminative representations can be highlighted. 
Second, our framework is endowed with the rotational invariance to some extent since the local adjacent relation of views is quite robust to rotation transformation. This holds for both aligned and unaligned 3D shapes. However, RNN based methods may be sensitive to the predefined viewpoints and thus cannot ensure the rotational invariance of the learned features, as suggested in~\cite{chen2018veram}.

% \textcolor{blue}{Even for unaligned 3D shapes, n-GLU is still capable of modeling the spatial dependency of view images for 3D shape since it models the local visual n-grams, while the RNN based method can not ensure the rotational invariance of the learned features, t herefore sensitive to the predefined view points, as suggested in~\cite{chen2018veram}.} 

\subsection{Attentional feature aggregation}
\label{sec:attentional feature aggregation}
N-gram learning unit outputs a new sequence of visual n-gram features with local spatial information incorporated. And for the downstream retrieval task, we need to aggregate them into a compact and discriminative representation. 

\begin{figure}[tb]
\includegraphics[width=0.45\textwidth]{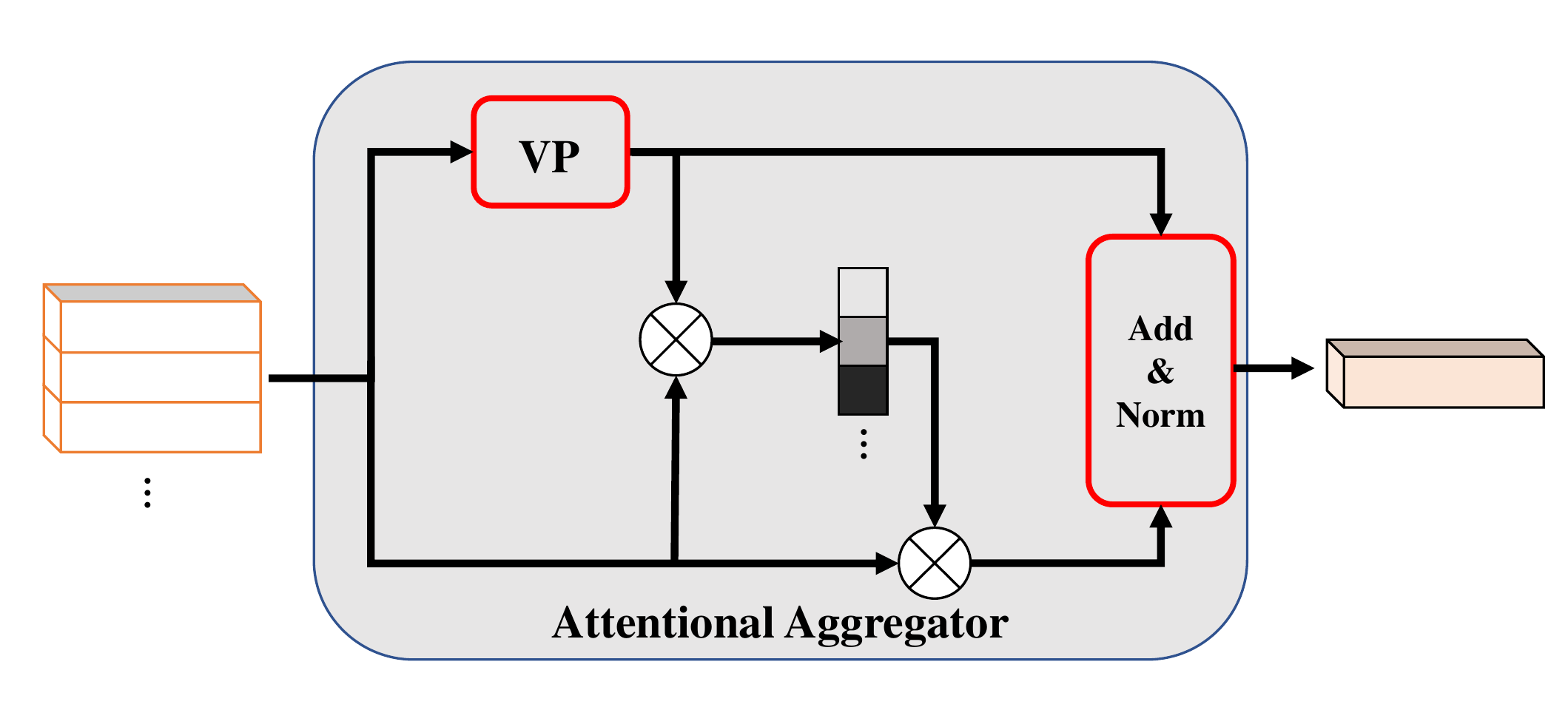}
\caption{The detailed structure of our proposed attention aggregator. Here VP represents view-pooling operation (\ie~max-pooling), which outputs global n-gram feature. The attention aggregator assigns weights to different n-gram features according to their similarities to the proxy global n-gram feature. A compact descriptor is obtained by combining all the n-gram features in a weighted sum manner.}
\label{fig:att_aggregator}
\vspace{-2ex}
\end{figure}

Max-pooling is a simple yet effective strategy. However, it may lead to sub-optimal performance as much information is lost by only keeping the maximum value. For better utilization of different n-gram view features and the reduction of information loss, we further propose to use attentional feature aggregation mechanism, which can be seen as a variant of self-attention module~\cite{vaswani2017attention}. The main difference is that we estimate the correlation between the global feature and each n-gram feature instead of all pairs of n-gram features. Besides, our attention module is free of parameters. The pipeline of our attention feature aggregation is presented in Fig.~\ref{fig:att_aggregator}, which can be formulated as 
 \begin{align}
\begin{split}\label{eq:att score}
 \beta_j  & = \frac{\exp(\phi_j(\mathbf{G_j}, \mathbf{g}_p))}{\sum_{i=1}^{|V|+1-n}\exp(\phi_i(\mathbf{G_i}, \mathbf{g}_p))}
\end{split}\\
\begin{split}\label{eq:2}
 \phi_j  &= \frac{\mathbf{G_j} * \mathbf{g}_p}{\sqrt{D'}}  
\end{split}\\
\begin{split}\label{eq:comb}
\mathbf{g}_{a} &= \sum_{j=1}^{|V|+1-n} \beta_j \mathbf{G}_{j}
\end{split}
\end{align}
where $\mathbf{g}_p$ and $\mathbf{g}_a$ denote the global representation for the sequence (via max pooling operation over $\mathbf{G}$) and the attentional output, respectively. $\mathbf{G_j}$ represents the $j$-th row of matrix $\mathbf{G}$~(\ie, $j$-th visual n-gram feature) and $\beta_j$ is the corresponding attention score. $\phi$ denotes the inner product with scale normalization to avoid extremely small gradients when the result is in large magnitude as suggested in \cite{vaswani2017attention}.
We elaborate the details of our attention aggregation mechanism below, which includes three main steps.

\vspace{1ex}\noindent\textbf{Estimation of attention score.}~We first adopt max-pooling operation over the set of the visual n-gram features $\mathbf{G}$ to obtain a global descriptor $\mathbf{g}_p$. We then assign attention scores to different n-gram features by estimating their correlations to the global descriptor $\mathbf{g}_p$ in an inner product manner, as formulated in Eq.~\eqref{eq:att score} and Eq.~\eqref{eq:2}.

\vspace{1ex}\noindent\textbf{View aggregation.}~With the estimated attention scores, the n-gram view features are aggregated into a compact representation $\mathbf{g}_{a}$ in a weighted sum fashion (see Eq.~\eqref{eq:comb}). In this way, different n-gram views, which contain different local spatial information, are effectively combined.

\vspace{1ex}\noindent\textbf{Residual connection.}~Similar to self-attention~\cite{vaswani2017attention}, we further introduce a residual connection to add the global descriptor $\mathbf{g}_p$ back to the aggregated feature $\mathbf{g}_{a}$ to reduce information loss, which is then normalized by a layer normalization operation~\cite{ba2016layer}. % We take the normalization result as the final representation of 3D shapes.

Note that the full pipeline of our attentional feature aggregation does not involve any learnable parameters. Therefore, it is very efficient.

\subsection{Remarks} 
\label{sec:multi scale n-gram fusion}

\noindent\textbf{Multi-scale n-gram feature fusion.}~Varying the n-gram size, n-GLU module can capture the spatial information in different scales (or the number of consecutive view images) and learn discriminative patterns of 3D shapes in distinct granularities. Therefore, we propose to employ multiple parallel branches for comprehensively characterizing the 3D shape from different scales. Each branch performs n-gram feature learning and attentional aggregation separately with different n-gram sizes. We then combine the aggregated features from all the branches in a concatenation manner. In this way, spatial information among the view images is better exploited. As presented in Fig.~\ref{fig:fig_arch}, we use three n-gram branches with n-gram sizes of 3, 5 and 7 in our VNN framework.

\vspace{1ex}\noindent\textbf{Training details.}~For the final recognition stage, we employ a simple network which is composed of only two fully connected layers. The first layer takes the combined feature as input and maps it into a lower-dimensional vector $\mathbf{g}_r \in \mathbb{R}^{512}$. The last layer predicts the category distribution based on $\mathbf{g}_r$. In our experiments, we adopt \textit{softmax} loss as the training objective. 

%-------------------------------------------------------------------------
\section{Experiments} 
\label{sec:exp}
%-------------------------------------------------------------------------
\subsection{Datasets}
In our experiments, we evaluate the proposed method on three public 3D shape retrieval datasets,~\ie,~ModelNet40~\cite{wu20153d}, ModelNet10~\cite{wu20153d} and ShapeNetCore55~\cite{chang2015shapenet}. ModelNet40 and ModelNet10 datasets are two subsets of the Princeton ModelNet dataset which contains 151,128 3D shapes divided into 660 categories. The ShapeNetCore55 dataset is a large-scale dataset from ShapeNet. 

The ModelNet40 contains 12,311 shapes from 40 common categories. In our experiments, we use the same training/test splits as popular MVCNN~\cite{su2015multi} and 3D ShapeNets~\cite{wu20153d}. The ModelNet10 dataset consists of 4,899 3D models in total from 10 classes. The training and testing sets contain 3,991 and 908 models, respectively. 

The ShapeNetCore55 dataset is introduced in the SHape REtrieval Contest (SHREC) 2016 competition track. It is a large-scale dataset, which is composed of 51,190 3D shapes from 55 shape categories. Each model in the dataset is also attached with a fine-grained subcategory from 204 subcategories in addition to the label from the 55 categories. Among these 3D shape data, 35,765 3D shapes (70\%) are provided for training and another 5,159 3D shapes (10\%) are for validation. The remaining 10,266 shapes (20\%) form the testing set. The dataset has two versions,~\ie,~``normal'' and ``perturbed'' versions. For the ``normal'' version, the 3D shapes are aligned. For the ``perturbed'' version, each 3D shape is arbitrarily oriented. Hence, the latter version is more challenging. To test the robustness of our method, we conduct experiments on both versions. 

\subsection{Evaluation metrics}

%-------------------------------------------------------------------------
In our experiments, we adopt five common metrics to evaluate the retrieval performance of our method against state-of-the-art methods, which are defined as follows: 
\begin{itemize}
 \setlength\itemsep{-0.5ex}
    \item Precision-Recall (PR) curve is used to visualize the retrieval performance. 
    \item Mean Average Precision (mAP) of the PR curve is used to give a quantitative evaluation.
    \item Area Under Curve (AUC) is the mean area under the PR curve. 
    \item F-Measure is the harmonic mean of recall and precision. 
    \item Normalized Discounted Cumulative Gain (NDCG) is a measure that assigns more weights to the relevant results at the top of the ranked list. 
\end{itemize}

Among these metrics, we use the PR curve, mAP, AUC on ModelNet40 and mAP, AUC on ModelNet10 to evaluate the retrieval performance. For ShapeNetCore55 dataset, F-Measure, mAP, and NDCG are adopted. 

\subsection{Implementation details}
We render a set of 2D greyscale images of size $224\times224$ for each 3D object, following the same rendering protocol as MVCNN~\cite{su2015multi}. For the aligned datasets,~\ie,~ModelNet40, ModelNet10 and ShapeNetCore55 ``normal'' datasets, we place virtual cameras around the 3D object every 30 degrees, obtaining 12 view images. For the unaligned ShapeNetCore55 ``perturbed'' dataset, 80 view images are rendered by placing virtual cameras at 20 vertices of the icosahedron and taking 4 views per camera using 0, 90, 180, and 270 degrees in-plane rotations.

During training, we adopt stochastic gradient descent (SGD) for optimization with momentum of 0.9 and weight decay of 0.0001. The learning rate is set to 0.001. We clip the gradients into the range [-0.01, 0.01] for training stability. And we train the model for 150 epochs with a mini-batch size of 8. In the inference time, we extract the output of the penultimate layer of our network, which is 512-dimensional, as the descriptor for each 3D object. 

We implement our method using PyTorch~\cite{paszke2017automatic}, and all the experiments are conducted on a server with eight NVIDIA Titan-X GPUs, an Intel i7 CPU and 64GB RAM. 
\subsection{Comparison with state-of-the-art methods}
To validate the effectiveness of our method, we first conduct experiments on two common 3D shape datasets-ModelNet40 and ModelNet10, in which 3D objects are assumed to be aligned. We then further conduct experiments on the more challenging ShapeNetCore55 dataset, which includes two versions,~\ie,~the ``normal'' version with aligned 3D objects and the ``perturbed'' version where the orientations of 3D shapes remain unknown. 

\vspace{1ex}\noindent\textbf{Comparison on ModelNet40.}~The comparison with state-of-the-art methods is listed in Table~\ref{tab:comparsions on modelnet}. 
We present the results of three representative model-based methods including SPH~\cite{kazhdan2003rotation}, 3DShapeNet~\cite{wu20153d} and DLAN~\cite{furuya2016deep}, and the results of several representative view-based methods including LFD~\cite{chen2003visual}, DeepPano~\cite{shi2015deeppano}, GIFT~\cite{bai2016gift}, MVCNN~\cite{su2015multi},  GVCNN~\cite{feng2018gvcnn}, RED~\cite{bai2017ensemble}, TCL~\cite{he2018triplet} and SeqViews~\cite{han2019SeqViews2SeqLabels} for extensive comparison. 
Moreover, we also re-implement MVCNN by inserting max-pooling operation at the fc-9 layer of VGG-A with batch normalization, as our baseline.
It can be observed that our method achieves very competitive performance, reaching 89.6\% in AUC and 88.9\% in mAP, outperforming most existing methods.
Concretely, our method surpasses the best model-based method DLAN by 3.9\% in terms of mAP. 
And when compared with other view-based methods, VNN outperforms GIFT, GVCNN, and TCL by 7.0\%, 3.2\%, and 0.9\% in mAP, respectively. 
It should be noted that GVCNN employs a stronger backbone GoogLeNet. Compared with RED, which is a sophisticated similarity fusion method, we outperform it by 2.6\% in mAP. Besides, our method achieves comparable performance with SeqViews for mAP~(88.9\% \textit{vs.} 89.1\%) while SeqViews leverages a stronger backbone VGG-19 for view feature extraction. To make a fair comparison, we further use VGG-19 for the experiment and achieve slightly better performance (89.3\% \textit{vs.} 89.1\% in mAP). When compared with our baseline, VNN obtains significant improvements, \eg, 89.6\% \textit{vs.} 73.7\% for AUC and 88.9\% \textit{vs.} 72.9\% for mAP. 
The consistent gains over the baseline and state-of-the-art methods demonstrate the superiority of our method.

% formats 
\begin{table}[!tb]
\small
\centering
\begin{tabular}{|l|*{2}{p{0.94cm}<{\centering}}|*{2}{p{0.94cm}<{\centering}}|}
\hline
\multirow{2}{*}{Methods} & \multicolumn{2}{c|}{ModelNet40} & \multicolumn{2}{c|}{ModelNet10}  \\
\cline{2-3} \cline{4-5} 
                         & AUC & mAP & AUC & mAP \\
\hline
\hline
SPH~\cite{kazhdan2003rotation} &34.5 &33.3 &46.0 &44.1\\
3DShapeNet~\cite{wu20153d} &49.9 &49.2 &69.3 &68.3  \\
DLAN~\cite{furuya2016deep} &- &85.0 &-  &90.6\\ \hline
LFD~\cite{chen2003visual} &42.0 &40.9 &51.7 &49.8 \\
DeepPano~\cite{shi2015deeppano} &77.6&76.8 &85.5&84.2\\
GIFT~\cite{bai2016gift}$^{\text{V-S}}$ &83.1 &81.9 &92.4 &91.1 \\
MVCNN~\cite{su2015multi}$^{\text{V-M}}$ &- &70.1 &- &-\\
MVCNN*~\cite{su2015multi}$^{\text{V-M}}$ &-  &80.2 &-  &-\\
RED~\cite{bai2017ensemble}$^\text{R-50}$ & 87.0 & 86.3 & 93.2  &92.2   \\
GVCNN~\cite{feng2018gvcnn}$^{\text{G}}$ &- &85.7 &- &-\\
TCL~\cite{he2018triplet}$^{\text{V-A}}$ &89.0 &88.0 &-  &-   \\
SeqViews~\cite{han2019SeqViews2SeqLabels}$^{\tiny\text{V-19}}$ &- & 89.1 &- &91.4\\\hline
MVCNN$^{\dagger}$$^{\text{V-A}}$ &73.7 &72.9 &80.8 &80.1 \\
\mycolor{Ours$^{\text{V-A}}$} &\mycolor{89.6} & \mycolor{88.9} & \mycolor{\textbf{93.5}} & \mycolor{\textbf{92.8}}  \\ 
\mycolor{Ours$^{\text{V-19}}$} &\mycolor{\textbf{90.2}} & \mycolor{\textbf{89.3}} & \mycolor{\textbf{-}} & \mycolor{\textbf{-}}  \\ \hline
\end{tabular}
\vspace{1ex}\caption{The comparison with state-of-the-art methods on ModelNet40 and ModelNet10. On the top are results of model-based methods. Results of view-based methods are listed in the middle. * means employing metric learning. $\dagger$ represents reproduced MVCNN result which is conducted under the same setting~(backbone network and pooling position) of the proposed method as the baseline. V-S, V-M, V-A, V-19, G and R-50 represent using VGG-S, VGG-M, VGG-A,  VGG-19, GoogLeNet and ResNet-50 architectures, respectively. }
\label{tab:comparsions on modelnet}
\vspace{-2ex}
\end{table}

\setlength{\tabcolsep}{8.7pt}
\begin{table*}[!tb]
\small
\centering
\begin{tabular}{|l|l|ccc|ccc|ccc|}
\hline
\multirow{2}{*}{ShapeNetCore55} & \multirow{2}{*}{Methods} & \multicolumn{3}{c|}{microALL} & \multicolumn{3}{c|}{macroALL} & \multicolumn{3}{c|}{microALL + macroALL}  \\
\cline{3-11}
& & F1 & mAP & NDCG & F1 & mAP & NDCG & F1 & mAP & NDCG  \\
\hline
\hline
\multirow{6}{*}{\textbf{normal}} & Wang~\cite{savva2016shrec16} & 39.1 &  82.3 & 88.6  & 28.6 & 66.1 & 82.0 &  33.8 & 74.2 & 85.3  \\  
& Li~\cite{savva2016shrec16}   &  58.2 & 82.9 & 90.4  & 20.1& 71.1 & 84.6 & 39.2 & 77.0 & 87.5 \\ 
& MVCNN~\cite{su2015multi}$^{\text{V-M}}$ & 76.4 & 87.3 & 89.9  & 57.5 & 81.7 & 88.0 & 66.9 & 84.5 &  89.0\\
& GIFT~\cite{bai2016gift}$^{\text{V-S}}$  & 68.9 & 82.5 & 89.6 &45.4 & 74.0 & 85.0 & 57.2 & 78.3 & 87.3 \\ 
& Kd-network~\cite{klokov2017escape}  & 74.3 & 85.0 & 90.5 & 51.9 & 74.6 & 86.4  & 63.1 & 79.8 & 88.5 \\
& \mycolor{Ours}$^{\text{V-A}}$ &  \mycolor{\textbf{78.9}} & \mycolor{\textbf{90.3}} & \mycolor{\textbf{92.8}}  & \mycolor{\textbf{61.4}} & \mycolor{\textbf{85.2}} & \mycolor{\textbf{91.7}}& \mycolor{\textbf{70.2}} & \mycolor{\textbf{87.8}} & \mycolor{\textbf{92.3}} \\
\hline 
\hline 
\multirow{7}{*}{\textbf{perturbed}} & Wang~\cite{savva2016shrec16} & 24.6 &  60.0 & 77.6  & 16.3 & 47.8 & 69.5 & 20.5 &  53.9 & 73.6\\ 
                     & Li~\cite{savva2016shrec16}   &  53.4 & 74.9 & 86.5  & 18.2 & 57.9 & 76.7 & 35.8 & 66.4 & 81.6\\ 
                    & MVCNN~\cite{su2015multi}$^{\text{V-M}}$ & 61.2 & 73.4 & 84.3  & 41.6 & 66.2 & 79.3 & 51.4 & 69.8 &  81.8\\
                     & GIFT~\cite{bai2016gift}$^{\text{V-S}}$  & 66.1 & 81.1 & 88.9  & 42.3 & 73.0 & 84.3 & 54.2 & 77.0 &  86.6\\
                     & Kd-network~\cite{klokov2017escape}  & 45.1 & 61.7 & 81.4 & 24.1 & 48.4 & 72.6  & 34.6 & 55.1 & 77.0\\
                     & TCL~\cite{he2018triplet}$^{\text{V-A}}$ & 67.9 & 84.0 & 89.5 & 43.9 & \textbf{78.3} & \textbf{86.9} & 55.9 & \textbf{81.2} & 88.2\\
                     &  \mycolor{Ours}$^{\text{V-A}}$ & \mycolor{\textbf{71.3}} & \mycolor{\textbf{84.3}} & \mycolor{\textbf{89.7}} & \mycolor{\textbf{50.1}} & \mycolor{78.0} & \mycolor{86.8} & \mycolor{\textbf{60.7}} & \mycolor{\textbf{81.2}} & \mycolor{\textbf{88.3}} \\
\hline
\end{tabular}
\vspace{1ex}\caption{The performance (\%) comparison on the ShapeNetCore55 dataset. On the top is the performance comparison on the ``normal'' version dataset, while comparison results on ``perturbed'' version are listed in the bottom rows.}
\label{shrec_table_results}
\vspace{-2ex}
\end{table*}

\begin{figure}[tb]
\centering
\includegraphics[width=0.45\textwidth]{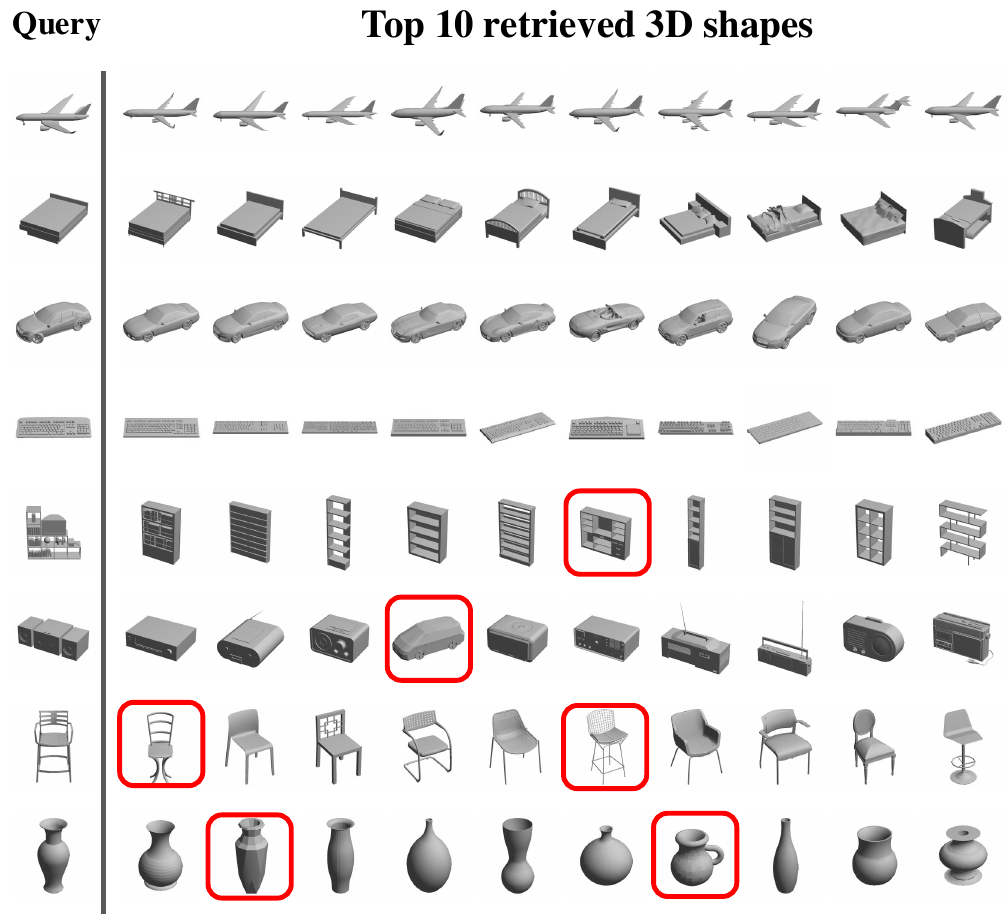}
\caption{Illustration of the retrieval examples on the ModelNet40 dataset. The query shapes are put on the first left column, and the top 10 retrieved shapes are on the right side. The retrieved false positives are highlighted by red boxes.}
\label{fig:retrieval_examples_modelnet40}
\vspace{-2ex}
\end{figure}

We present several retrieval examples on the ModelNet40 dataset in Fig.~\ref{fig:retrieval_examples_modelnet40}. It can be seen that our method can retrieve highly relevant 3D objects for the queries. Note that the retrieved false positives also share similar shapes with the query, \eg,~the vase in the last row.

\vspace{1ex}\noindent\textbf{Comparison on ModelNet10.} We present the results on the Modelnet10 dataset in Table~\ref{tab:comparsions on modelnet}. As shown, our method yields an AUC of 93.5\% and an mAP of 92.8\%, considerably outperforming state-of-the-art methods. Comparing with DLAN~\cite{furuya2016deep}, which is a superior model-based method utilizing rich rotation-invariant 3D local features, we achieve an improvement of 2.2\% in terms of mAP. In addition, VNN outperforms DeepPano~\cite{shi2015deeppano}, GIFT~\cite{bai2016gift} and RED~\cite{bai2017ensemble} by 8.6\%, 1.7\%, and 0.6\% in terms of mAP, respectively. The comparison with SeqViews~\cite{han2019SeqViews2SeqLabels} is especially valuable since SeqView leverages a stronger backbone network for view feature extraction. Nevertheless, the proposed method surpasses it by 1.4\% in terms of mAP. We also provide the reproduced result of MVCNN under the same settings with VNN as a baseline. We can observe that VNN significantly boosts the mAP of MCNNN by 12.7\%, suggesting the importance of local spatial relation for 3D object retrieval and the effectiveness of the proposed method.

\vspace{1ex}\noindent\textbf{Comparison on ShapeNetCore55 dataset.} We conduct experiments on both the ``normal'' and the ``perturbed'' versions of the ShapeNetCore55 dataset.
For the convenience of comparison with other methods, we provide the average value for the two kinds of evaluation metrics~(\ie,~microALL and macroALL). 
The comparison for both versions is listed in Table~\ref{shrec_table_results}. 
As shown, significant improvements over state-of-the-art methods are achieved on both versions of the dataset. In particular, on the ``normal'' version, when compared with Kd-network~\cite{klokov2017escape}, which is model-based, we improve the mAP by 9\%. And compared with MVCNN~\cite{su2015multi}, improvements of over 3\% in terms of all the evaluation metrics are obtained. At the same time, the comparison results on the more challenging ``perturbed'' version also prove the effectiveness of our method. 
When compared with TCL~\cite{he2018triplet}, which adopts an end-to-end metric learning loss function, we achieve comparable performance in terms of both mAP and NDCG.
However, for F1, nearly 5\% improvements are obtained. It demonstrates that the proposed method can be naturally extended to the cases where orientations of 3D shapes are agnostic.

%---------------------------------------------------------------------------------
\subsection{Ablation study}

In this section, we present ablation experiments of the proposed method on the ModelNet40 dataset. Concretely, we will study the effects of n-gram size, combinations of different n-gram sizes, attentional aggregation mechanism and its complementarity with metric learning methods.

\vspace{1ex}\noindent\textbf{Effect of n-gram size.} we first explore the effect of n-gram size on the model performance. The n-gram size is a very important hyper-parameter because it affects the context window for computing the n-gram features. We set the n-gram size to be 1, 3, 5, 7 and study their impacts on ModelNet40. The comparison results are reported in Table~\ref{table:ablation on n-gram size}. As shown, when the n-gram size is 1, which means adopting uni-gram without considering the context of each view images, we can only reach an mAP of 80.7\%. However, when we increase the n-gram size to 3, we see an improvement of nearly 4\%. It suggests that more discriminative representations can be obtained by incorporating local spatial information via the visual n-grams (n\textgreater1). The best result is obtained when we set the n-gram size to be 5, which reaches 88.0\% and 87.3\% in terms of AUC and mAP, respectively. When the n-gram size is greater than 5, we see some degree of degradation on retrieval performance. 

\setlength{\tabcolsep}{7.8pt}
\begin{table}[!tb]
\small
\centering
\begin{tabular}{|l|*{2}{p{1.0cm}<{\centering}}|*{2}{p{1.0cm}<{\centering}}|}
\hline
\multirow{2}{*}{N-gram size} & \multicolumn{2}{c|}{Max-Pooling} & \multicolumn{2}{c|}{With Attention}  \\
\cline{2-3} \cline{4-5} 
                         & AUC & mAP & AUC & mAP \\
\hline
\hline
 1 & 80.3 & 79.5 & 81.5 & 80.7 \\
 3 & 84.0 & 83.2 & 85.6 & 84.9 \\
 5 & 85.5 & 84.7 & 88.0 & 87.3 \\
 7 & 85.1 & 84.3 & 86.6 & 85.9 \\
 3+5 & 88.3 & 87.6 & 88.7 & 88.0 \\
3+7 & 87.8 & 87.0 & 88.3 & 87.5 \\
 5+7 & 87.9 & 87.1 & 88.9 & 88.1 \\
 3+5+7 & \textbf{88.9} & \textbf{88.2} & \textbf{89.6} & \textbf{88.9} \\ \hline 
\end{tabular}
\vspace{1ex}\caption{Ablation analysis on the ModelNet40 dataset.}
\label{table:ablation on n-gram size}
\vspace{-2ex}
\end{table}

\begin{figure}[tb]
\centering
\includegraphics[width=0.45\textwidth]{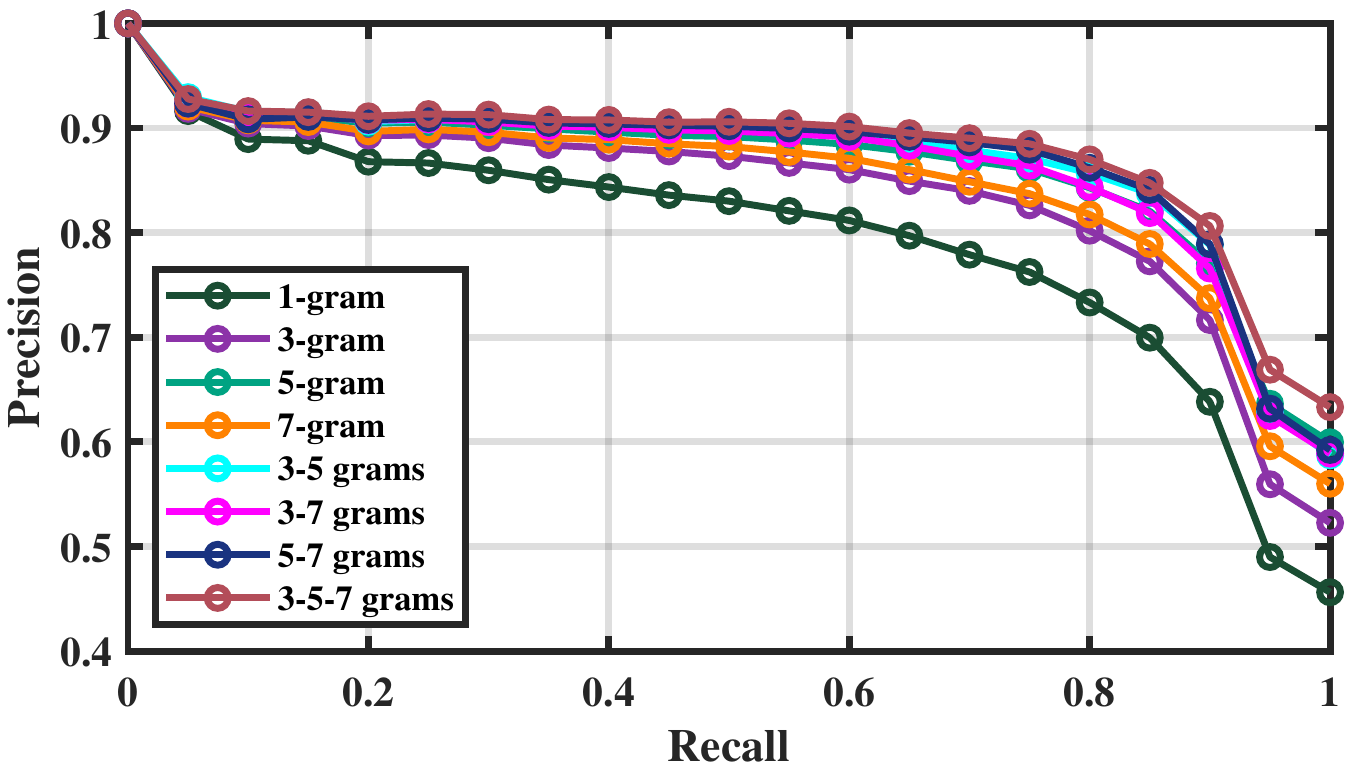}
\caption{PR curves on the ModelNet40 dataset for different settings of n-gram size.}
\label{fig:pr_modelnet40}
\vspace{-2ex}
\end{figure}

\vspace{1ex}\noindent\textbf{Effect of fusion on different n-grams.} We further discuss how the combinations of different n-gram sizes influence the retrieval performance. Concretely, we enumerate all the combinations of three n-gram sizes~(\ie,~3-gram, 5-gram and 7-gram). As shown in Table~\ref{table:ablation on n-gram size}, combining different kinds of gram sizes yields consistent improvements over only utilizing a single gram size, suggesting different n-gram modules are complementary to each other and combining them can effectively improve the retrieval performance. \textcolor{black}{Besides, it should be noted that different combinations 
in Table~\ref{table:ablation on n-gram size} share the same retrieval efficiency since they output representations of the same dimension~(\ie~512)}. 

The PR curves are shown in Fig.~\ref{fig:pr_modelnet40}, which intuitively demonstrate the retrieval performance based on each n-gram size and their different combinations. 

\vspace{1ex}\noindent\textbf{Effect of the attentional aggregation mechanism.} We compare the attentional aggregation mechanism with the widely adopted max-pooling operation under different settings for n-gram size. As presented in Table~\ref{table:ablation on n-gram size}, the attentional aggregation mechanism demonstrates consistent improvements over max-pooling. It should be noted that our attentional aggregation mechanism is parameter-free, suggesting it can serve as a more powerful alternative to max-pooling for better utilization of multi-view features.

\vspace{1ex}\noindent\textbf{Complementary to metric learning.} Triplet-Center Loss (TCL)~\cite{he2016deep} is an effective metric learning loss and has achieved superior performance on multiple 3D shape benchmarks. Hence, we further combine the softmax loss with the triplet-center loss to validate the complementarity of our method with metric learning method. As shown in Table~\ref{table:combine_with_tcl}, TCL yields an improvement of AUC 0.9\% and mAP of 0.6\%, demonstrating the potential of our method when combined with existing metric learning methods. % PR curves are presented in Fig.~\ref{fig:pr_modelnet40_tcl} to visualize the retrieval performance more comprehensively.
\setlength{\tabcolsep}{23.3pt}
\begin{table}[!tb]
\small
\centering
\begin{tabular}{|l|*{2}{p{0.6cm}<{\centering}}|}
\hline
\multirow{1}{*}{Methods} &  AUC & mAP \\
\cline{1-2} 
\hline
\hline 
\multicolumn{1}{|l|}{Ours} & 89.6 & 88.9 \\
\multicolumn{1}{|l|}{Ours + TCL~\cite{he2018triplet}}  & \textbf{90.5} & \textbf{89.5} \\
\hline
\end{tabular}
\vspace{1ex}\caption{The performance (\%) of our method with and without TCL on ModelNet40.}
\label{table:combine_with_tcl}
\vspace{-2ex}
\end{table}
% should we comment this 
%\begin{figure}[tb]
%\centering
%\includegraphics[width=0.9\linewidth]{./Figures/modelnet_40_pr_tcl.pdf}
%\caption{PR curves on the ModelNet40 dataset with and without TCL loss.}
%\label{fig:pr_modelnet40_tcl}
%\end{figure}
\section{Conclusion}
\label{sec:conclusion}
In this paper, we propose a novel framework named View N-gram Network~(VNN) to model the spatial relationships of multi-view images for 3D objects, which can learn discriminative representations for 3D object retrieval task. The core component of VNN is the n-Gram Learning Unit~(n-GLU), which first divides multiple view images into a set of visual n-grams efficiently and then learns the intra-gram feature effectively. In this way, local spatial information is leveraged. Attentional aggregation mechanism is adopted over the learned n-gram features. Moreover, we propose to fuse representations under different n-gram sizes. Experimental results on multiple 3D shape benchmarks demonstrate the superiority of the learned 3D shape representations from the proposed method. In the future, we would like to explore other alternatives to the concatenation method for better fusing the multi-scale n-gram features. 
\paragraph{Acknowledgements:} This work was supported by NSFC 61573160, to Dr. Xiang Bai by the National Program for Support of Top-notch Young Professionals and the Program for HUST Academic Frontier Youth Team 2017QYTD08.

% original  \bibliographystyle{ieee}
{\small
\bibliographystyle{ieee_fullname}
\bibliography{egbib}
}

\end{document}